\documentclass{article}

\usepackage[nonatbib,final]{neurips_2022}

\usepackage{graphicx}

\usepackage[utf8]{inputenc} % allow utf-8 input
\usepackage[T1]{fontenc}    % use 8-bit T1 fonts
\usepackage{hyperref}       % hyperlinks
\usepackage{url}            % simple URL typesetting
\usepackage{booktabs}       % professional-quality tables
\usepackage{amsfonts}       % blackboard math symbols
\usepackage{nicefrac}       % compact symbols for 1/2, etc.
\usepackage{microtype}      % microtypography
\usepackage{xcolor}         % colors
\usepackage{subcaption}
\usepackage{lipsum}
\usepackage{cleveref}

\title{Towards Transformer-based Homogenization of Satellite Imagery for Landsat-8 and Sentinel-2}

\author{
  Venkatesh Thirugnana Sambandham\\
  Department of Computer Science\\
  Otto-von-Guericke University Magdeburg\\
  \texttt{venkatesh.thirugnana@st.ovgu.de} \\
   \And
   Konstantin Kirchheim \\
   Department of Computer Science \\
   Otto-von-Guericke University Magdeburg\\
   \texttt{konstantin.kirchheim@ovgu.de} \\
    \And
    Sayan Mukhopadhaya \\ 
    BASF Digital Farming GmbH \\ 
   \texttt{sayan.mukhopadhaya@xarvio.com} \\ 
   \And
   Frank Ortmeier \\
   Department of Computer Science \\
   Otto-von-Guericke University Magdeburg\\
   \texttt{ortmeier@ovgu.de} \\
}

\begin{document}

\maketitle

\begin{abstract}
Landsat-8 (NASA) and Sentinel-2 (ESA) are two prominent multi-spectral imaging satellite projects that provide publicly available data.
The multi-spectral imaging sensors of the satellites capture images of the earth's surface in the visible and infrared region of the electromagnetic spectrum.
Since the majority of the earth's surface is constantly covered with clouds, which are not transparent at these wavelengths, many images do not provide much information.
To increase the temporal availability of cloud-free images of a certain area, one can combine the observations from multiple sources.
However, the sensors of satellites might differ in their properties, making the images incompatible. 
This work provides a first glance at the possibility of using a transformer-based model to reduce the spectral and spatial differences between observations from both satellite projects.
We compare the results to a model based on a fully convolutional UNet architecture. 
Somewhat surprisingly, we find that, while deep models outperform classical approaches, the UNet significantly outperforms the transformer in our experiments. 
\end{abstract}

\section{Introduction}
\begin{figure}
\centering
\begin{subfigure}{0.32\textwidth}
    \includegraphics[width=\textwidth]{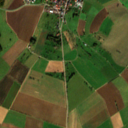}
    \caption{Sentinel-2}
    \label{fig:sentinel-example}
\end{subfigure}
\hfill
\begin{subfigure}{0.32\textwidth}
    \includegraphics[width=\textwidth]{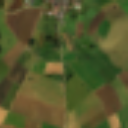}
    \caption{Landsat-8}
    \label{fig:landsat-example}
\end{subfigure}
\hfill
\begin{subfigure}{0.32\textwidth}
    \includegraphics[width=\textwidth]{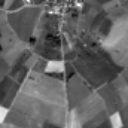}
    \caption{Panchromatic}
    \label{fig:pan-chromatic-example}
\end{subfigure}
\hfill
\caption{A comparison of Sentinel-2 and Landsat-8 RGB composites of the same region, as well as the corresponding panchromatic band from Landsat-8. We see that the resolution of the Sentinel-2 is higher than the resolution of the Landsat-8 image.}
\label{fig:satellite-comparison}
\end{figure}

A large number of satellite clusters are constantly monitoring the physical characteristics of the earth's surface using active and passive electromagnetic sensors.
These satellites capture data across multiple bands of the electromagnetic spectrum, which can be utilized in several important applications, such as crop health monitoring~\cite{roy2014landsat,mukhopadhaya2016land}. 
For these applications, achieving a high temporal resolution for a particular area of the earth's surface is advantageous. 
Landsat-8 and Sentinel-2 are two prominent earth observation satellite missions that provide publicly available data. 
A comprehensive study on the revisiting time of Landsat-8, 9 as well as Sentinel-2 A and B found that a combination of all 4 sensors could potentially reduce the duration between two revisits to 2.3 days on average~\cite{LS89S2Combine}.
However, not all of the available images are usable due to occlusions caused by clouds, which are not transparent at the corresponding wavelengths.
Furthermore, the sensors have different properties and consequentially, several additional challenges have to be considered:
\begin{enumerate}
    \item The spectral differences caused by different characteristics of the  imaging sensors used in each project have to be minimized.
    \item The differences in the spatial resolution of the image sources have to be handled.
\end{enumerate}
For example, consider \cref{fig:satellite-comparison}: loosely speaking, our goal is to generate image with the quality of \cref{fig:sentinel-example}  by a combination of \cref{fig:landsat-example} and the panchromatic band in \cref{fig:pan-chromatic-example}.

Approaches based on Deep Learning~\cite{schmidhuber2015deep,lecun2015deep} play a key role in several applications in the field of remote sensing ~\cite{RSDL}. 
Extensive research was conducted on the topic of deep learning based super-resolution (SR), with some techniques showing exceptional improvements in the perceptual quality of low-resolution images~\cite{SR_SURVEY}. 
In particular, the UNet~\cite{UNet}, which is based on an encoder-decoder architecture, provides good results for super-resolution~\cite{UNetSR}.

In this work, we explore the use of vision transformers for the harmonization of images from Landsat-8 and Sentinel-2. 
We hypothesize that transformers can outperform UNet-based models in the considered setting, due to the ability of the incorporated attention mechanism to account for complex, non-local interaction between variables. 

\section{Background}
Reducing the spectral differences between the images acquired from multiple sensors is a well-researched topic~\cite{spectral_adjustments,SRF_21,L8_RED_EDGE}. 
Most previous works derive an explicit compensation function to convert the reflectances of one image source to another.
    
Deep Learning based approaches have recently been applied to image super-resolution and re-construction applications~\cite{SRCNN, DNCNN}.
Generally, deeper Convolutional Neural Network (CNN) architectures (with residual connections~\cite{ResNet} and feature concatenation~\cite{densenet}) yield better results in common computer vision tasks.
Deep Learning based super-resolution architectures were also applied to enhance the spatial resolution of low-resolution multi-spectral images~\cite{S2RE,PSS2,s2ll8}.

Motivated by their superior performance on Natural Language Processing tasks, transformer-based models have recently been adapted for Computer Vision tasks.
Vision transformers (ViT) like the SwinIR~\cite{swinir} and the ESRT~\cite{ESRT} have successfully been applied to image reconstruction and super-resolution.

\section{Models and Dataset}
\subsection{Data}
We collect pairs of images from both satellite sources that were taken over the Area Of Interest (AOI) on the same day. 
The AOI considered in this work covers a large region of Germany. 
The areas used for training, validation, and testing are disjoint from each other to avoid target leakage. 
The high-resolution (HR) images from Sentinel-2 act as ground truth images; the objective of the model is to homogenize the common bands from the low-resolution (LR) images from Landsat-8 with respect to the HR source.

\subsection{Preprocessing}
After we fetch data from both sources, the images from various bands are converted to reflectances, filtered for cloud coverage, and then corrected for inter-sensor misregistration issues.
The images are finally broken down into smaller patches of size $1.28$ km$\times1.28$ km. 
Each patch consists of 7-channels which include multi-spectral bands Blue, Green, Red, Near Infrared (NIR), two Short Wave Infrared (SWIR), and the panchromatic image from the LR source.
After patching there were 63,740 patches for training, 22,344 patches for testing, and 11,908 patches for validation.
The LR images are bicubically upsampled to the size of HR images to act as the input to the models.

\subsection{Neural Network Models and Training}
In this study, we compare two deep super-resolution architectures: a UNet~\cite{UNet,UNetSR} and a slightly modified Efficient Super Resolution Transformer (ESRT)~\cite{ESRT}.
The pixel-shuffle upsampling block in the ESRT architecture was removed, since the images were already upsampled to the target size in the pre-processing step.
We train the models for 50 epochs, using the Adam optimizer with a mini-batch size of 20, minimizing the $\ell_1$ norm between the ground truth and the prediction. 
We use $1\times 10^{-5}$ as our learning rate.
Each model was trained on a single 32 GB NVIDIA V100 GPU. 
Since the GPU could not hold 20 patches at a time while training the transformer-based model, we used gradient accumulation over four batches of size five.

\section{Results}
\begin{figure}
\centering
\begin{subfigure}{0.49\textwidth}
    \includegraphics[width=\textwidth]{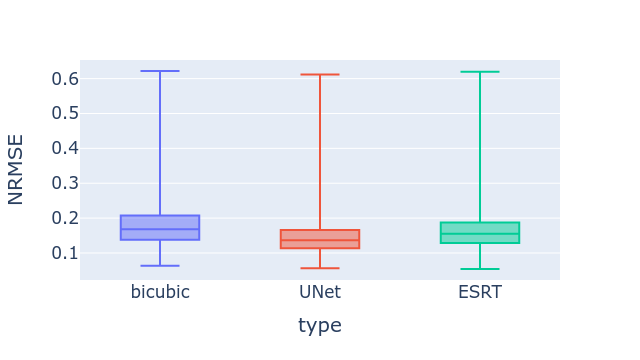}
    \caption{NRMSE}
    \label{fig:ESRT_distribution}
\end{subfigure}
\hfill
\begin{subfigure}{0.49\textwidth}
    \includegraphics[width=\textwidth]{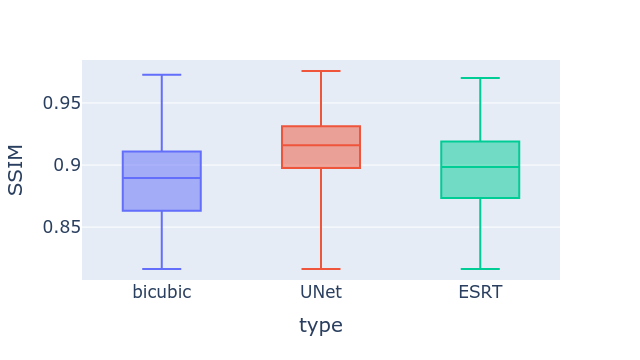}
    \caption{SSIM}
    \label{fig:UNet_distribution}
\end{subfigure}
% \hfill
\caption{Distribution of metrics calculated for different methods on the test set.}
\label{fig:metrics}
\end{figure}

We evaluate the models using a pixel-wise Normalized Root Mean Squared Error (NRMSE) and perception-based Structural Similarity Index Measure (SSIM) as described in \cref{tab:metrics}.
The distribution of the calculated metrics from the test data set is given as box plots in \cref{fig:metrics}.
In the quantitative evaluation, we observe that the UNet significantly outperformed the baseline upsampling and the ESRT architecture with a mean SSIM of 0.91.
Although the transformer outperformed the baseline, it is not on par with the UNet in terms of both metrics. 

\begin{figure}
    \centering
    \includegraphics[width=\textwidth]{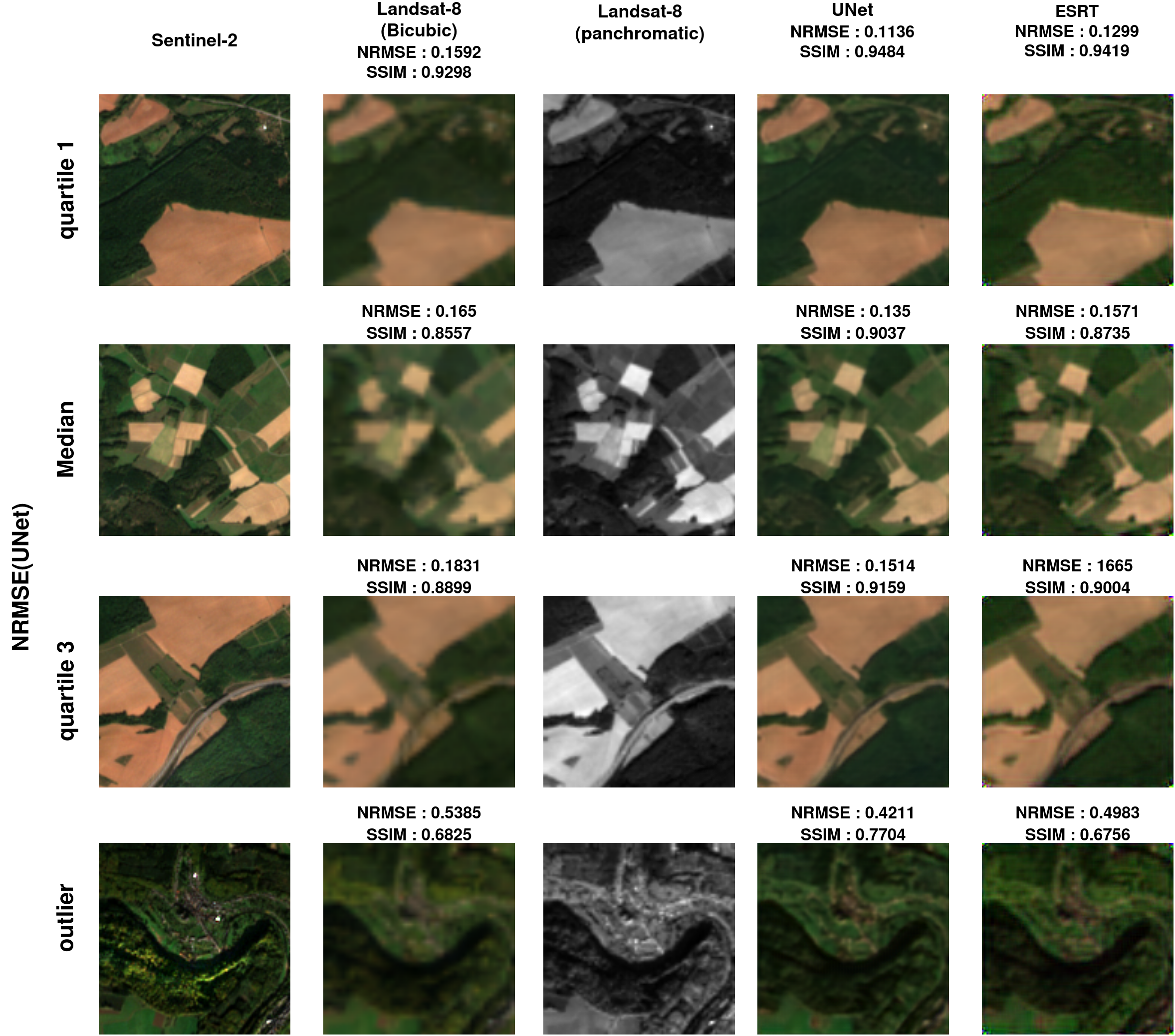}
    \caption{Results from sample patches across different levels of UNet NRMSE metrics over the test set. Outliers often contain high-frequency components that are not present in the panchromatic band.}
    \label{fig:EVAL}
\end{figure}

A qualitative comparison of the RGB composites of samples at different performance levels (as measured by the UNets NRMSE) is given in \cref{fig:EVAL}.
Significant improvement in the spatial quality can be observed in the images generated by both neural networks.
However, some artifacts appear in the images generated by the ESRT, especially in the corners of the patches.
Another unsurprising observation is that models fail to reconstruct high-frequency components of the image, like settlements, whose presence can cause significant drops in performance metrics.

\begin{table}[h]
\centering
\caption{Mean and standard deviation of performance metrics on the test set.}
\label{tab:metrics}
\vspace{1mm}
\begin{tabular}{|l|l|l|}
\hline
        & \multicolumn{1}{c|}{SSIM} & \multicolumn{1}{c|}{NRMSE} \\ \hline
Bicubic & 0.8763 $\pm$ 0.043            &  0.1839 $\pm$ 0.056          \\ \hline
UNet    & \textbf{0.9114 $\pm$ 0.028}   & \textbf{0.1460 $\pm$ 0.048}    \\ \hline
ESRT    & 0.8898 $\pm$ 0.039            &    0.1705 $\pm$ 0.064          \\ \hline
\end{tabular}
\end{table}
 
\section{Conclusion}
We presented preliminary results for a transformer-based approach for harmonizing multi-spectral images of the earths surface provided by the Landsat-8 and Sentinel-2 satellites. 
The model constitutes a component in a larger pipeline that enables higher temporal resolution of the observations of a specific region, which can be utilized in several downstream applications, for example, crop health monitoring. 
In first experiments, we somewhat surprisingly find that, while deep models outperform classical approaches like bilinear interpolation, the UNet outperforms the more complex transformer. 

A possible explanation for the superior performance of the UNet is that our initial hypothesis does not hold: non-local interactions might not be as relevant for this application, and the UNets translation invariance, coupled with a sufficiently large receptive field (i.e., sufficient depth), constitutes a better inductive bias for the considered task.

\bibliographystyle{plain}
\bibliography{main.bib}
\end{document}